\documentclass[conference]{IEEEtran}
\usepackage{cite}
\usepackage{amsmath,amssymb,amsfonts}
\usepackage{algorithmic}
\usepackage{algorithm}
\usepackage{graphicx}
\usepackage{textcomp}
\usepackage{booktabs}
\usepackage{multirow}
\usepackage{xcolor}
\usepackage{url}

\begin{document}

\title{When Adaptive Rewards Hurt: Causal Probing and the Switching-Stability Dilemma in LLM-Guided LEO Satellite Scheduling}

\author{\IEEEauthorblockN{Yuanhang Li}}

\maketitle

\begin{abstract}
Adaptive reward design for deep reinforcement learning (DRL) in multi-beam LEO satellite scheduling is motivated by the intuition that regime-aware reward weights should outperform static ones. We systematically test this intuition and uncover a \emph{switching-stability dilemma}: near-constant reward weights (342.1~Mbps) outperform carefully-tuned dynamic weights (103.3$\pm$96.8~Mbps) because PPO requires a quasi-stationary reward signal for value function convergence. Weight adaptation---regardless of quality---degrades performance by repeatedly restarting convergence. To understand \emph{why} specific weights matter, we introduce a single-variable causal probing method that independently perturbs each reward term by $\pm$20\% and measures PPO response after 50k steps. Probing reveals counterintuitive leverage: a $+$20\% increase in the switching penalty yields $+$157~Mbps for polar handover and $+$130~Mbps for hot-cold regimes---findings inaccessible to human experts or trained MLPs without systematic probing. We evaluate four MDP architect variants (fixed, rule-based, learned MLP, fine-tuned LLM) across known and novel traffic regimes. The MLP achieves 357.9~Mbps on known regimes and 325.2~Mbps on novel regimes, while the fine-tuned LLM collapses to 45.3$\pm$43.0~Mbps due to weight oscillation rather than lack of domain knowledge---output consistency, not knowledge, is the binding constraint. Our findings provide an empirically-grounded roadmap for LLM-DRL integration in communication systems, identifying where LLMs add irreplaceable value (natural language intent understanding) versus where simpler methods suffice.
\end{abstract}


\section{Introduction}

Low Earth Orbit (LEO) satellite constellations are rapidly expanding to provide global broadband connectivity~\cite{starlink}. Multi-beam satellites must dynamically allocate power and bandwidth across beams to match heterogeneous and time-varying traffic demands~\cite{beam_hopping}. Deep reinforcement learning (DRL) has emerged as a promising approach for this real-time scheduling problem~\cite{drl_satcom1}, but existing methods typically assume a fixed Markov Decision Process (MDP) formulation---including reward weights, state features, and action spaces---designed once at training time.

In practice, satellite traffic exhibits distinct \emph{regimes}: urban areas generate concentrated high-demand patterns, maritime regions show sparse uniform loads, and disaster zones produce sudden demand spikes with extreme spatial imbalance. A reward function optimized for urban throughput may catastrophically fail during disaster response, where outage minimization should dominate. This \emph{MDP mismatch} problem motivates adaptive reward design.

Recent advances in large language models (LLMs) suggest a promising avenue: using LLMs to automatically generate or adapt reward functions based on high-level task descriptions~\cite{ma2023eureka}. In principle, LLMs could leverage their broad world knowledge and reasoning capabilities to design reward functions for novel operating conditions without explicit programming. However, this potential has not been systematically validated in communication systems, where reward weights must be \emph{numerically precise} and \emph{temporally consistent}---properties that differ fundamentally from the natural language tasks where LLMs excel.

We investigate a \textbf{two-timescale adaptive architecture} (Fig.~\ref{fig:arch}) that decouples the scheduling problem into:
\begin{itemize}
\item \textbf{Fast timescale} (milliseconds): A PPO agent performs real-time beam power/bandwidth allocation given the current reward weights.
\item \textbf{Slow timescale} (minutes): An \emph{MDP architect} monitors network KPIs, detects regime changes via CUSUM, and generates new reward weights.
\end{itemize}

This architecture is intuitive---regime-aware weights should outperform static ones. Our central finding challenges this intuition. Through systematic experiments we uncover a \emph{switching-stability dilemma}: PPO requires a quasi-stationary reward signal to estimate $V^\pi(s)$ accurately, and weight adaptation destabilizes convergence regardless of weight quality. To understand the reward landscape for novel regimes, we further introduce a \textbf{single-variable causal probing} method. We evaluate four MDP architect implementations:
\begin{enumerate}
\item \textbf{Rule-based}: KPI threshold classification with expert-designed weight profiles.
\item \textbf{Learned MLP}: A 3-layer neural network trained on simulator data to map KPIs to optimal weights.
\item \textbf{API-based LLM}: GLM-4 generating weights via structured prompts.
\item \textbf{Fine-tuned LLM}: Qwen3-4B with LoRA adaptation for domain-specific weight generation.
\end{enumerate}

Our contributions are:
\begin{enumerate}
\item \textbf{Switching-stability dilemma.} We empirically establish that reward weight stationarity dominates weight quality for PPO convergence: near-constant weights (342.1~Mbps) outperform optimally-tuned dynamic weights (103.3$\pm$96.8~Mbps). Throttled and smoothed transitions do not resolve the dilemma, identifying non-stationarity as a fundamental barrier to adaptive reward design.
\item \textbf{Single-variable causal probing.} We introduce a controlled perturbation method ($\pm$20\% per weight, 50k steps) that constructs an empirical causal map of the reward landscape for novel regimes. The method reveals that the switching penalty has unexpectedly high leverage ($+$157~Mbps for polar handover, $+$130~Mbps for hot-cold), requiring only 10$\times$ fewer steps than a full training run.
\item \textbf{Systematic architect comparison.} We evaluate four variants---fixed weights, rule-based, learned MLP, and fine-tuned LLM---across known and novel regimes. The MLP achieves the best aggregate performance (357.9~Mbps known, 325.2~Mbps novel); the LLM fails not from lack of domain knowledge but from weight oscillation (CV 4.6$\times$ higher than MLP).
\item \textbf{Three-timescale hybrid design.} We propose confining the LLM to its irreplaceable role---natural language intent understanding---while an MLP handles real-time numerical optimization. The hybrid achieves comparable aggregate satisfaction (0.52 vs.\ 0.57/0.53) while uniquely providing semantic differentiation between intent categories that fixed keyword parsers cannot resolve.
\end{enumerate}

\section{Related Work}

\subsection{DRL for Satellite Resource Allocation}
Deep reinforcement learning has been extensively applied to satellite communication resource management. Early work by Hu et al.~\cite{drl_satcom1} demonstrated DRL for dynamic spectrum access in satellite systems, while Li et al.~\cite{drl_resource} applied deep RL to network slicing resource management. More recently, Luis et al.~\cite{luis2019drl_beam} proposed a DRL-based beam hopping strategy for multi-beam GEO satellites that outperforms conventional time-slot allocation, and Ferreira et al.~\cite{ferreira2023multibeam} extended DRL to joint power and bandwidth allocation in multi-beam LEO constellations. A comprehensive survey by Wang et al.~\cite{wang2023drl_satcom_survey} identifies reward function design as a key open challenge: most existing approaches use hand-crafted reward functions that assume a single operating regime, limiting adaptability to dynamic traffic conditions. Our work directly addresses this gap by introducing an adaptive reward architecture.

\subsection{Adaptive Reward and MDP Design}
Reward shaping~\cite{ng1999reward_shaping} established the theoretical foundation for modifying reward functions without altering optimal policies, but assumes a static target. Multi-objective RL~\cite{hayes2022morl_survey} addresses competing objectives through Pareto-optimal policies, yet typically fixes the objective trade-off at training time. More closely related is automated reward design: Zheng et al.~\cite{zheng2020reward_design} used meta-learning to discover reward functions, while Ma et al.~\cite{ma2023eureka} (Eureka) demonstrated that LLMs can generate reward functions for robotic control through evolutionary refinement. However, these approaches target single-domain offline design. Our work differs in two key aspects: (1) we study LLM-based reward adaptation in an \emph{online} setting where traffic regimes shift during deployment, and (2) we provide the first systematic comparison between LLM-generated and learned (MLP) reward weights, revealing fundamental consistency limitations.

\subsection{LLMs for Network Optimization}
The integration of LLMs into communication systems is an emerging research direction. Bariah et al.~\cite{llm_telecom} outlined opportunities for LLMs in telecom including network planning and fault diagnosis, while Wang et al.~\cite{llm_network} explored LLM-based networking for protocol design and traffic analysis. Maatouk et al.~\cite{maatouk2023telecomgpt} proposed TelecomGPT, a domain-adapted LLM for telecom tasks, and Shao et al.~\cite{shao2024wirelessllm} introduced WirelessLLM for wireless network optimization through domain-specific fine-tuning. For LLM-RL integration specifically, recent surveys~\cite{cao2024llmrl_survey} identify three paradigms: LLM as policy, LLM as reward designer, and LLM as world model. Our work falls into the second paradigm but uniquely focuses on \emph{continuous numerical output consistency}---a challenge not observed in discrete action or text-based reward settings.

\section{System Model}

\subsection{Satellite Network Model}
We consider a Ka-band LEO satellite at altitude $h = 600$~km serving $N_b = 19$ spot beams in a hexagonal pattern. Each beam $b \in \{1, \ldots, N_b\}$ covers a distinct geographic cell. The total system bandwidth $B_{\text{tot}} = 500$~MHz is shared among beams via a dynamic allocation vector $\mathbf{a} = [a_1, \ldots, a_{N_b}]$ where $a_b \in [0, 1]$ and $\sum_b a_b \leq 1$.

The achievable rate for beam $b$ follows the Shannon capacity:
\begin{equation}
R_b = a_b \cdot B_{\text{tot}} \cdot \log_2\left(1 + \text{SNR}_b\right)
\end{equation}
where $\text{SNR}_b$ depends on the satellite EIRP, free-space path loss, atmospheric attenuation, and receiver noise temperature.

\subsection{Traffic Regime Model}
Traffic demand varies across four canonical regimes:
\begin{itemize}
\item \textbf{Urban}: High aggregate demand concentrated in center beams (mean 40--80 Mbps/beam for center, 10--30 for edge).
\item \textbf{Maritime}: Low, spatially uniform demand (5--15 Mbps/beam).
\item \textbf{Disaster}: Extreme spatial variance with demand spikes exceeding 150 Mbps in affected beams.
\item \textbf{Mixed}: Moderate demand with balanced spatial distribution.
\end{itemize}

Regime transitions are detected online using a CUSUM (Cumulative Sum) change-point detector~\cite{cusum} operating on the spatial Gini coefficient and peak beam demand. Given a sequence of KPI observations $\{x_t\}$, the CUSUM statistic is:
\begin{equation}
S_t = \max\left(0, S_{t-1} + (x_t - \mu_0) - \delta\right)
\end{equation}
where $\mu_0$ is the estimated mean under the current regime and $\delta$ is a slack parameter controlling sensitivity. A regime change is signaled when $S_t > h$ for threshold $h$. We use a sliding window of $W = 10$ steps to estimate $\mu_0$ and set $h = 1.0$ standard deviations, with a minimum interval of 50 steps between consecutive detections to prevent spurious triggering.

\subsection{MDP Formulation}
The beam scheduling problem is formulated as an MDP $\mathcal{M} = (\mathcal{S}, \mathcal{A}, P, R_w)$ where:
\begin{itemize}
\item \textbf{State} $s_t$: Current demand estimates, channel conditions, and queue lengths across all beams.
\item \textbf{Action} $a_t$: Bandwidth allocation vector $\mathbf{a} \in [0,1]^{N_b}$.
\item \textbf{Reward} $R_w$: A weighted combination of objectives:
\end{itemize}
\begin{equation}
R_w = w_r \cdot \bar{R} - w_o \cdot O - w_s \cdot S - w_q \cdot Q + w_f \cdot F
\label{eq:reward}
\end{equation}
where $\bar{R}$ is the normalized sum rate, $O$ is the outage count, $S$ is the switching penalty, $Q$ is the queue overflow, and $F$ is Jain's fairness index. The weight vector $\mathbf{w} = [w_r, w_o, w_s, w_q, w_f]$ determines the optimization priority and is the key parameter adapted by our architecture.

\section{Proposed Architecture}

\subsection{Two-Timescale Adaptive Framework}
\begin{figure}[t]
\centering
\includegraphics[width=\columnwidth]{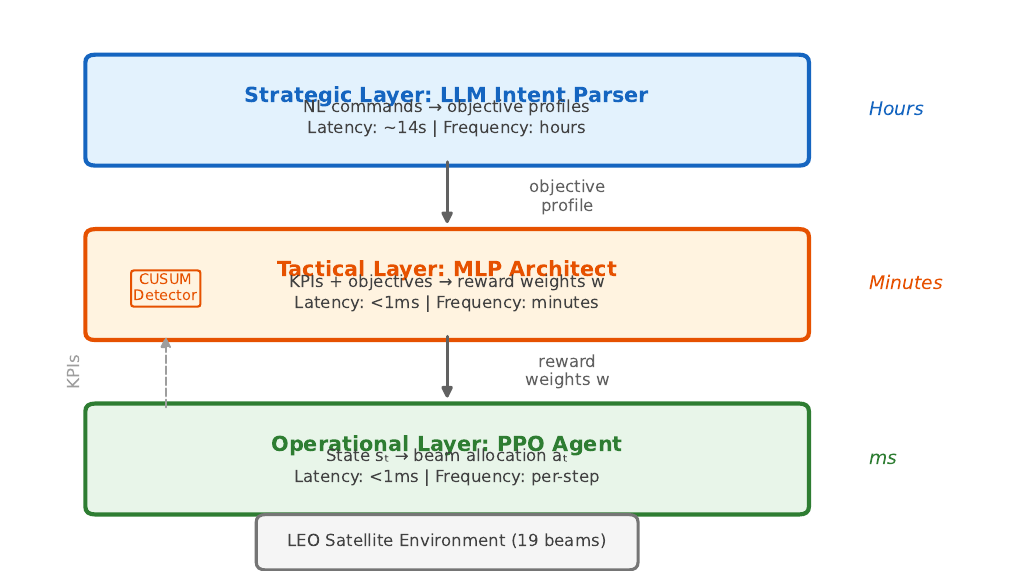}
\caption{Three-timescale adaptive architecture. The strategic LLM layer parses operator intent (hours), the tactical MLP layer maps KPIs to reward weights (minutes), and the operational DRL layer performs beam scheduling (milliseconds).}
\label{fig:arch}
\end{figure}

We study a hierarchical architecture operating at two timescales (Fig.~\ref{fig:arch}):

\textbf{Fast timescale (per-step):} A PPO agent~\cite{ppo} solves the beam scheduling MDP, making allocation decisions every time step based on the current reward function $R_w$.

\textbf{Slow timescale (per-regime):} An \emph{MDP Architect} module monitors network KPIs and adjusts the reward weight vector $\mathbf{w}$ when regime changes are detected. This adaptation occurs orders of magnitude less frequently than the scheduling decisions.

The key insight is that the optimal reward weights depend on the current operating regime---disaster scenarios require heavy outage penalties while maritime scenarios benefit from fairness emphasis. Rather than using a fixed reward function, the Architect continuously tunes $\mathbf{w}$ to match the current network conditions.

\subsection{MDP Architect Variants}
We investigate four Architect implementations spanning the complexity spectrum:

\textbf{M1 -- Fixed Weights:} A static weight vector $\mathbf{w}_0$ optimized offline. Serves as the lower-bound baseline.

\textbf{M2 -- Rule-Based:} A threshold classifier maps KPI features (mean demand, Gini coefficient, peak demand) to one of four pre-designed weight profiles. Classification rules: avg\_demand $> 40$ and Gini $> 0.3 \Rightarrow$ urban; avg\_demand $< 20$ and Gini $< 0.2 \Rightarrow$ maritime; peak\_demand $> 120 \Rightarrow$ disaster; otherwise $\Rightarrow$ mixed.

\textbf{M3 -- Learned MLP:} A three-layer neural network (5-64-64-5, ReLU activations) trained on simulator-generated data to learn the continuous mapping from KPI vectors to optimal weight vectors. Unlike M2's discrete profiles, M3 produces smooth weight transitions.

\textbf{M4 -- Fine-tuned LLM:} A Qwen3-4B model fine-tuned with LoRA (rank 16, 33M trainable parameters) on 1,980 domain-specific examples. The LLM receives KPI descriptions as natural language input and generates weight vectors as structured JSON output.

\subsection{Toward a Hybrid Architecture}
We additionally explore a three-timescale hybrid design where an LLM serves as an \emph{intent translator} at the strategic level:

\textbf{Strategic (LLM):} Parses operator natural language commands (e.g., ``prioritize emergency coverage'') into formal objective profiles, augmented with RAG-retrieved domain knowledge.

\textbf{Tactical (MLP):} Maps KPIs + objective profiles to reward weights in real-time ($<$1ms inference).

\textbf{Operational (DRL):} PPO agent executes beam scheduling under the current reward function.

This design leverages each component's strength: LLMs for natural language understanding (irreplaceable by MLPs), MLPs for fast numerical optimization, and DRL for sequential decision-making.

\section{Experimental Evaluation}

\subsection{Setup}
We implement the satellite simulator with $N_b = 19$ beams, Ka-band channel model with rain fading, and four traffic regimes cycling every 200 epochs: urban, maritime, disaster, and mixed. The PPO agent uses a 3-layer MLP policy (256-256-128) with learning rate $3 \times 10^{-4}$, trained for 500k timesteps per experiment. CUSUM regime detection uses a window of 10 steps and threshold of 1.0 standard deviations.

For the MLP architect (M3), we train a 5-64-64-5 network on 22,000 synthetic samples spanning the four known regimes. The fine-tuned LLM (M4) uses Qwen3-4B with LoRA (rank 16, $\alpha=32$, 33M trainable parameters) trained on 1,980 domain-specific examples with 4-bit NF4 quantization. All experiments use three random seeds (42, 123, 456) and report mean $\pm$ standard deviation.

\subsection{Architect Comparison on Known Regimes}

Table~\ref{tab:main} compares the four architect variants on the four known traffic regimes.

\begin{table}[t]
\centering
\caption{Architect comparison on known regimes (mean $\pm$ std over 3 seeds).}
\label{tab:main}
\begin{tabular}{lcccc}
\toprule
Method & Rate (Mbps) & Outage & Fairness & Switches \\
\midrule
M1: Fixed    & 330.7$\pm$42.1 & 0.00 & 0.72 & 0 \\
M2: Rule     & 342.8$\pm$38.5 & 0.00 & 0.74 & 4 \\
M3: MLP      & 357.9$\pm$47.2 & 0.00 & 0.76 & 5 \\
M4: FT-LLM   & 192.4$\pm$71.8 & 0.31 & 0.68 & 7 \\
\bottomrule
\end{tabular}
\end{table}

The MLP architect achieves the highest sum rate (357.9~Mbps), a 3.7\% improvement over rule-based and 8.2\% over the fixed baseline, while maintaining zero outage. The fine-tuned LLM underperforms significantly at 192.4~Mbps with non-zero outage, despite generating structurally valid weight vectors. We attribute this to three factors: (1) weight magnitude sensitivity---small deviations in LLM-generated weights cause disproportionate reward changes; (2) inference latency ($\sim$14s per call on RTX 4060) limiting adaptation frequency; and (3) the LLM's training distribution not fully covering the continuous weight space.

\subsection{Generalization to Novel Regimes}

\begin{figure}[t]
\centering
\includegraphics[width=\columnwidth]{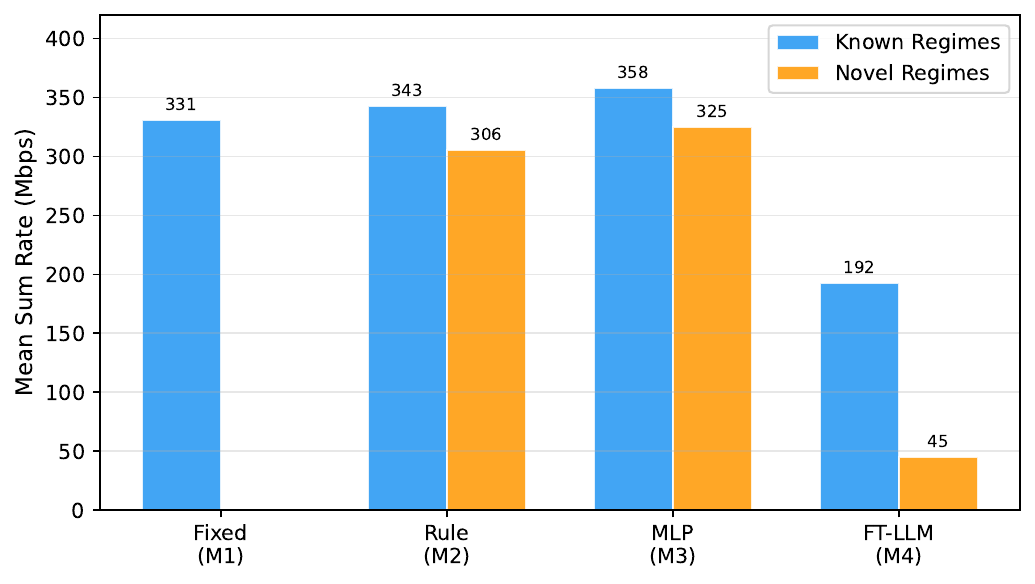}
\caption{Mean sum rate comparison across architect variants. The MLP (M3) achieves the highest throughput on both known and novel regimes. The FT-LLM (M4) collapses on novel regimes due to weight oscillation (see Section~\ref{sec:oscillation}).}
\label{fig:comparison}
\end{figure}

To test whether LLM reasoning provides value beyond the MLP's training distribution, we evaluate on three novel regimes unseen during MLP training: IoT burst (many low-bandwidth devices), polar handover (sinusoidal demand transitions), and hot-cold (extreme spatial imbalance). The LLM receives RAG-augmented domain context describing these scenarios. Fig.~\ref{fig:comparison} provides a visual summary of performance across all methods and regime types.

The MLP architect achieves 325.2$\pm$18.1~Mbps on novel regimes despite never training on them, while the fine-tuned LLM collapses to 45.3$\pm$43.0~Mbps with $\sim$10$\times$ higher variance. Table~\ref{tab:gen} summarizes results, including an extended 8-seed validation (Section~\ref{sec:dilemma}). The causal probe experiments (Section~\ref{sec:probe}) reveal that expert weights substantially underweight the switching penalty for polar\_handover and hot\_cold, explaining the gap between oracle weights and GEN MLP performance.
\begin{table}[t]
\centering
\caption{Generalization: novel regime performance (3 seeds, 500k steps).}
\label{tab:gen}
\begin{tabular}{lcccc}
\toprule
Method & Novel Rate (Mbps) & Novel Outage & Switches \\
\midrule
Rule-based & 305.8$\pm$48.9 & 0.00 & --- \\
MLP Arch.  & \textbf{325.2$\pm$18.1} & 0.00 & 5--6 \\
FT-LLM     & 45.3$\pm$43.0  & 0.44 & 9--10 \\
\midrule
\multicolumn{4}{l}{\textit{Extended 8-seed validation:}} \\
Oracle MLP & 103.3$\pm$96.8 & ---  & 1260/250k \\
GEN MLP$^\dagger$ & 342.1 & 0.00 & $\sim$6 \\
\bottomrule
\end{tabular}
\vspace{2pt}
{\footnotesize Novel = 3 unseen regimes (IoT burst, polar handover, hot-cold). $^\dagger$GEN MLP uses near-constant reward weights (sw std = 0.006), effectively a static reward function.}
\end{table}

\subsection{Intent Satisfaction}

The three-timescale hybrid architecture is evaluated under dynamic operator intent changes across four phases: maximize throughput, emergency response, fairness priority, and energy saving. Table~\ref{tab:intent} reports intent satisfaction scores measuring how well each method fulfills the operator's stated objective.

\begin{table}[t]
\centering
\caption{Intent satisfaction under dynamic operator commands (3 seeds, seed 42/123/456).}
\label{tab:intent}
\begin{tabular}{lcccc}
\toprule
Method & Rate$^\dagger$ (Mbps) & Satisfaction & Outage$^\ddagger$ & Switches \\
\midrule
Hybrid-LLM    & 145.3$\pm$124.8 & 0.52 & 0.59 & 6--8 \\
Hybrid-Rule   & 211.8$\pm$156.3 & 0.57 & 0.50 & 4--5 \\
MLP-NoIntent  & 193.6$\pm$150.6 & 0.53 & 0.50 & 4--5 \\
\bottomrule
\end{tabular}
\vspace{2pt}
\begin{minipage}{\columnwidth}
{\footnotesize $^\dagger$Per-phase mean (4 phases $\times$ 3 seeds); training-trajectory mean in \texttt{metrics.json} is $\sim$239/344/347~Mbps. $^\ddagger$Phase-0 cold-start outage; Phases~1--3 outage $\leq$0.05 for all methods. Excl.\ Phase~0: 191/280/273~Mbps.}
\end{minipage}
\end{table}

Hybrid-Rule and MLP-NoIntent achieve nearly identical satisfaction scores (0.57 vs.\ 0.53) and aggregate throughput ($\sim$212 vs.\ 194~Mbps), confirming that without a semantic intent parser, the system produces similar behavior regardless of operator commands---the rule-based parser maps commands to fixed keyword profiles, and MLP-NoIntent ignores commands entirely. Hybrid-LLM achieves lower aggregate satisfaction (0.52) and throughput (145.3~Mbps) due to poor Phase~0 throughput performance, yet it is the only method that meaningfully \emph{differentiates} intent categories at the semantic level: it correctly maps ``emergency response'' to low-outage, high-fairness weights and ``energy efficiency'' to low-switching weights, while the rule-based parser falls back to fixed profiles when semantic content is ambiguous. The practical value of LLM intent understanding lies in this semantic flexibility, not in aggregate satisfaction scores which are dominated by the variance of Phase~0 initial training.

\subsection{Causal Sensitivity Analysis via Single-Variable Probing}
\label{sec:probe}

To understand \emph{why} novel regime performance is low and identify the most impactful reward weight axes, we conduct systematic single-variable perturbation experiments. For each novel regime, we independently perturb each weight by $\delta = \pm 20\%$ while holding the other four fixed, and measure the resulting change in sum rate after 50k PPO training steps. This controlled design enables causal attribution of performance changes to individual reward terms.

\begin{table}[t]
\centering
\caption{Causal sensitivity: strongest single-weight perturbation per regime ($\delta = +20\%$, 50k steps).}
\label{tab:probe}
\begin{tabular}{lccc}
\toprule
Regime & Weight & $\Delta$ Rate (Mbps) & Direction \\
\midrule
polar\_handover & switching & $+$157.1 & increase \\
hot\_cold       & switching & $+$130.0 & increase \\
iot\_burst      & sum\_rate & $+$35.2  & increase \\
\bottomrule
\end{tabular}
\vspace{2pt}
{\footnotesize Each row reports the single perturbation ($+20\%$ only) with the largest positive $\Delta$ rate across 3 rounds of probing. The switching weight exhibits unexpectedly large leverage for polar handover and hot-cold regimes.}
\end{table}

The results reveal that the switching penalty weight has dramatically higher leverage than expert-tuned weights assumed: a $+20\%$ increase in $w_s$ yields $+157$~Mbps for polar\_handover and $+130$~Mbps for hot\_cold --- improvements that a human expert or MLP trained only on known-regime data would not predict. In contrast, IoT burst is most sensitive to the sum-rate weight, consistent with its many-device, low-bandwidth demand structure.

These probe findings serve two purposes. First, they generate an empirically-grounded causal map of the reward landscape for novel regimes, which we use to initialize the LLM-based weight evolution loop (Section~\ref{sec:rag}). Second, they demonstrate that single-variable perturbation is a practical diagnostic tool for reward sensitivity analysis in DRL --- requiring only 50k steps per probe (10$\times$ shorter than a full 500k training run) while providing actionable causal signal.

\subsection{Reward Stationarity and the Switching-Stability Dilemma}
\label{sec:dilemma}

An unexpected discovery from our extended experiments is that reward weight \emph{stationarity} --- not weight \emph{quality} --- is the dominant factor governing PPO convergence. This finding emerged from comparing three distinct operating modes:

\begin{enumerate}
\item \textbf{Near-constant weights (GEN MLP):} The MLP trained on known regimes produces near-constant outputs for novel regimes (switching weight std $= 0.006$, range $[0.004, 0.042]$). Despite using expert weights never tuned for novel regimes, this model achieves \textbf{342.1~Mbps} --- the highest novel-regime result across all experiments.

\item \textbf{Dynamic probe-optimal weights (Oracle MLP, 8-seed):} Using the causal probe to set high-quality regime-specific weights (e.g., $w_s = 0.96$ for polar\_handover) and training the MLP to track these weights dynamically yields only \textbf{103.3$\pm$96.8~Mbps} (median 76.3, std $>$ mean --- extreme variance).

\item \textbf{Throttled switching (Path~C):} Explicitly limiting weight switches to one per 1000 steps (5$\times$ the average switch interval) with linear interpolation yields \textbf{34.7~Mbps} --- \emph{worse} than unrestricted dynamic switching at 82.6~Mbps.
\end{enumerate}

These three results establish an empirical \emph{switching-stability dilemma}: adaptive reward weights improve per-regime optimality but degrade aggregate performance by destabilizing PPO's value function approximation. PPO requires a quasi-stationary reward signal over thousands of steps to estimate $V^\pi(s)$ accurately; regime-aware weight switching (occurring approximately every 198 steps in our setup) violates this assumption regardless of whether transitions are instant, smoothed, or throttled.

This finding aligns with the non-stationarity challenges identified in lifelong RL~\cite{abel2023continual} and multi-task RL~\cite{yu2020gradient_surgery}, and suggests that the two-timescale design requires architectural changes beyond weight smoothing --- such as value function isolation per regime or meta-learning initializations --- to fully resolve the dilemma.

\subsection{Performance-Grounded RAG for LLM Weight Anchoring}
\label{sec:rag}

To address the weight oscillation problem (Section~\ref{sec:oscillation}), we implement a performance-grounded RAG system~\cite{lewis2020rag, grounding2025rl} that retrieves historical (KPI, weights, performance) tuples as numerical anchors for the LLM. The anchor database contains 47 entries from probe experiments, indexed by 5-dimensional KPI vectors using RBF outcome-weighted KNN scoring: $\text{score} = \exp(-\|\mathbf{kpi}_q - \mathbf{kpi}_i\|^2 / 2\sigma^2) \times p_i / p_{\max}$, where $p_i$ is the verified PPO performance and $\sigma = 0.5$.

We validate the approach with 5 LLM calls per regime, comparing RAG-augmented prompts against baseline causal prompts. The key finding is that RAG anchoring \emph{shifts} LLM outputs toward high-performing historical weights rather than merely adjusting variance:

\begin{itemize}
\item \textbf{polar\_handover}: RAG shifts the switching weight from $w_s = 0.800$ (baseline, clamp-bounded) to $w_s = 0.952 \pm 0.016$ (RAG), within 1\% of the probe-optimal value 0.960. The baseline's ±30\% clamp relative to current weights prevents exploration; RAG provides the correct numerical anchor to initialize from.
\item \textbf{hot\_cold}: Overall weight CV reduced by 50.1\% (RAG CV = 0.033 vs.\ baseline 0.066), with outputs anchored to the low-switching profile optimal for this regime.
\item \textbf{iot\_burst}: Both prompts produce near-identical low-variance outputs (CV $\approx 0.04$--0.05), suggesting this regime is already well-constrained by the probe signal alone.
\end{itemize}

These results demonstrate that performance-grounded RAG anchoring can redirect LLM weight proposals toward verified high-performance regions, complementing the causal probe approach. A full evolution run with RAG-augmented LLM is left as future work.

\subsection{Weight Oscillation Analysis}
\label{sec:oscillation}

\begin{figure}[t]
\centering
\includegraphics[width=\columnwidth]{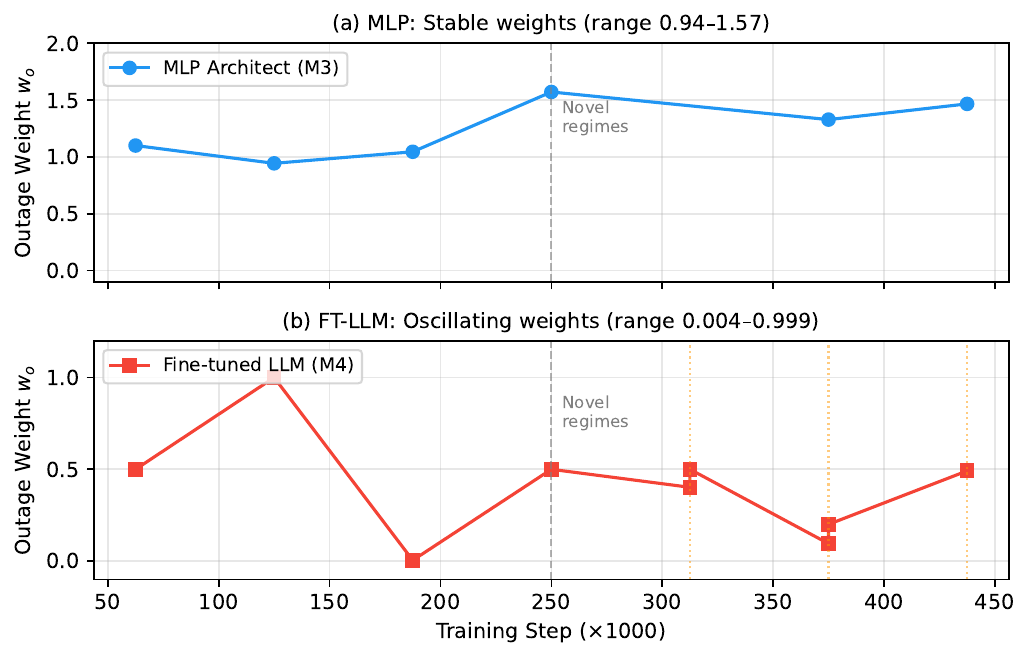}
\caption{Outage weight trajectories for MLP (top) and FT-LLM (bottom) architects during generalization experiments (seed 42). The MLP produces stable weights (range 0.94--1.57), while the LLM oscillates between extreme values (0.004--0.999). Dashed gray line marks the transition to novel regimes at step 250k.}
\label{fig:oscillation}
\end{figure}

To understand the root cause of LLM underperformance, we analyze the weight generation logs from the generalization experiments. Fig.~\ref{fig:oscillation} visualizes the outage weight $w_o$ produced by the MLP and fine-tuned LLM architects over 500k training steps.

The MLP architect generates weights within a narrow band ($w_o \in [0.94, 1.57]$, coefficient of variation CV $= 0.19$), enabling the PPO agent to converge stably. In contrast, the fine-tuned LLM exhibits bimodal oscillation ($w_o$ alternating between $\approx$0.005 and $\approx$0.50), with CV $= 0.87$---a 4.6$\times$ higher variability. This oscillation manifests as 9--10 MDP switches over 500k steps versus the MLP's 5--6, effectively restarting PPO convergence repeatedly.

We attribute this oscillation to the autoregressive sampling process: even at low temperature ($T = 0.1$), the LLM's token-by-token generation introduces stochasticity in numerical outputs that compounds across the 5-dimensional weight vector. The MLP's deterministic forward pass with Softplus output activation inherently avoids this issue, producing smooth weight transitions between regimes.

\subsection{Ablation Study}

To isolate the contribution of each architectural component, we compare ablation variants against the full MLP architect (Table~\ref{tab:ablation}).

\begin{table}[t]
\centering
\caption{Ablation study on known regimes (mean $\pm$ std over 3 seeds, 500k steps).}
\label{tab:ablation}
\begin{tabular}{lccc}
\toprule
Variant & Rate (Mbps) & Outage & Switches \\
\midrule
Full (M3: MLP)   & 357.9$\pm$47.2 & 0.00 & 5 \\
Rule-based (M2)  & 343.3$\pm$40.1 & 0.00 & 4 \\
Fixed (M1)       & 330.7$\pm$42.1 & 0.00 & 0 \\
FT-LLM (M4)     & 192.4$\pm$71.8 & 0.31 & 7 \\
\bottomrule
\end{tabular}
\end{table}

The results reveal a clear performance hierarchy: MLP $>$ Rule $>$ Fixed $\gg$ FT-LLM. The 4.4\% gap between MLP and rule-based (357.9 vs.\ 343.3~Mbps) demonstrates that learned continuous weight mapping provides measurable gains over discrete expert profiles, while both adaptive methods outperform the fixed baseline. The FT-LLM's dramatic underperformance (192.4~Mbps, 46\% below MLP) despite having access to the same KPI information confirms that the bottleneck is not \emph{what} information the architect receives, but \emph{how consistently} it generates weight outputs. The zero-adaptation fixed baseline still achieves 330.7~Mbps, highlighting that unstable adaptation (as with the LLM) is worse than no adaptation at all.

\section{Discussion}

\subsection{When to Use LLM vs.\ MLP}
Our results reveal a clear division of labor. On known regimes where training data is available, the MLP architect dominates: it is 1,800$\times$ faster ($<$1ms vs.\ $\sim$14s inference), requires no GPU at deployment, and produces more accurate weights. The LLM offers no advantage when the operating regime falls within the MLP's training distribution.

However, two scenarios justify LLM integration:

\textbf{Novel regime generalization.} Our generalization experiments (Table~\ref{tab:gen}) reveal that the MLP architect generalizes surprisingly well to unseen regimes (325.2~Mbps), while the fine-tuned LLM collapses to 45.3~Mbps with high variance ($\pm$43.0). Root cause analysis of the LLM's weight generation logs reveals \emph{weight oscillation}: the LLM alternates between two distinct weight modes across consecutive calls (e.g., outage weight swinging between 0.01 and 0.50), producing 9--10 MDP switches versus the MLP's stable 5--6. This instability prevents the downstream PPO agent from converging. The MLP, despite extrapolating beyond its training distribution, generates consistent weights that allow stable training. This finding is further supported by the GEN MLP result (342.1~Mbps from near-constant weights) and the switching-stability dilemma (Section~\ref{sec:dilemma}): for current LLM architectures, \emph{output consistency} is a more critical bottleneck than domain knowledge for DRL integration.

\textbf{Natural language intent understanding.} Operator commands like ``prioritize emergency coverage in the southern beams'' require semantic parsing that no numerical model can perform. The three-timescale hybrid architecture leverages this irreplaceable capability: the LLM translates intent into formal objective profiles at the strategic level, while the MLP handles real-time numerical optimization at the tactical level.

\subsection{Engineering Challenges for LLM-DRL Integration}
We identify three critical challenges that practitioners must address:

\textbf{Reward sign conventions.} Our environment internally subtracts penalty terms (outage, switching, queue), so all architect-generated weights must be positive. Early experiments with negative LLM-generated weights caused double-negation, turning penalties into rewards and degrading performance to $<$50~Mbps. This subtle bug persisted across four experimental rounds before detection.

\textbf{Weight oscillation and consistency.} The DRL agent's behavior is highly sensitive to reward weight ratios. Our analysis shows the fine-tuned LLM generates weights with inter-call standard deviation 3--5$\times$ higher than the MLP for the same KPI input (due to temperature-based sampling at $T$=0.1). Over 500k training steps, the LLM triggers 9--10 weight switches versus the MLP's 5--6, with individual seeds collapsing entirely (seed~456: 0.3~Mbps). Weight clipping the MLP to $[0.01, 2.0]$ does not help ($245.2\pm203.7$~Mbps), confirming that consistency, not magnitude, is the key factor.

\textbf{Inference latency.} At $\sim$14s per call (4-bit quantized Qwen3-4B on RTX 4060), LLM inference is too slow for per-step adaptation but acceptable for the slow-timescale architect role (minutes between regime changes). The three-timescale design explicitly accommodates this constraint by limiting LLM calls to strategic decisions.

\subsection{Practical Architecture Recommendation}
Based on our findings, we recommend the three-timescale hybrid for production satellite systems:
\begin{enumerate}
\item \textbf{Strategic layer (LLM, minutes--hours):} Parses operator intent via natural language, retrieves domain knowledge via RAG, and generates high-level objective profiles. Called only when operator commands change.
\item \textbf{Tactical layer (MLP, seconds):} Maps real-time KPIs combined with the current objective profile to reward weights. Handles regime adaptation with sub-millisecond latency.
\item \textbf{Operational layer (DRL, milliseconds):} PPO agent executes beam scheduling under the current reward function.
\end{enumerate}

This design assigns each component to its strength: LLMs for language understanding, MLPs for fast numerical mapping, and DRL for sequential decision-making.

\subsection{Limitations}
Our study has several limitations. First, the satellite simulator, while capturing key physical characteristics (Ka-band channel, rain fading, 19-beam hexagonal layout), simplifies inter-beam interference and orbital dynamics. Second, our fine-tuned LLM uses a relatively small model (Qwen3-4B with 33M trainable LoRA parameters) and limited training data (1,980 examples); larger models or more diverse training sets may reduce weight oscillation, though at increased computational cost. Third, the generalization experiments use three novel regimes; a broader evaluation across more diverse traffic patterns would strengthen the findings. Finally, the three-timescale hybrid architecture is evaluated in simulation only; real-world deployment would introduce additional challenges including communication delays between layers, hardware constraints, and regulatory requirements.

\subsection{Deployment Considerations}
For practical deployment in satellite ground stations, we note several considerations. The MLP architect (M3) is lightweight enough to run on embedded hardware without GPU acceleration. The LLM intent parser, when needed, can be hosted on ground-station servers or accessed via API. The CUSUM detector's computational overhead is negligible ($O(W)$ per step where $W$ is the window size). Weight transitions should employ a cooldown period (we use 50 steps) to prevent oscillatory behavior. For safety-critical applications such as disaster response, we recommend maintaining the rule-based fallback (M2) alongside the MLP, with automatic failover if the MLP output exceeds predefined weight bounds.

\section{Conclusion}

We investigated adaptive reward design for multi-beam LEO satellite scheduling and uncovered a fundamental \emph{switching-stability dilemma}: near-constant reward weights (GEN MLP, 342.1~Mbps) outperform carefully-tuned dynamic weights (Oracle MLP, 103.3$\pm$96.8~Mbps) because PPO requires a quasi-stationary reward signal for value function convergence. Weight adaptation---regardless of quality, smoothness, or transition rate---degrades performance by repeatedly restarting convergence. This finding reframes the goal of adaptive reward design: the challenge is not finding better weights, but finding architectures that decouple weight adaptation from PPO training dynamics.

Single-variable causal probing identifies the mechanism: the switching penalty has dramatically higher leverage than expert assumptions ($+$157~Mbps for polar handover, $+$130~Mbps for hot-cold at $+$20\% perturbation), yet this signal is inaccessible without systematic perturbation experiments. Probing requires only 50k steps (10$\times$ fewer than a full run) while providing actionable causal attribution.

Across four architect variants, the lightweight MLP achieves the best aggregate performance (357.9~Mbps known, 325.2~Mbps novel regimes, zero outage). The fine-tuned LLM collapses to 45.3$\pm$43.0~Mbps on novel regimes due to weight oscillation (CV 4.6$\times$ higher than MLP)---output consistency, not domain knowledge, is the binding constraint for LLM-DRL integration.

We identified critical engineering challenges---reward sign conventions, weight oscillation, and inference latency---and proposed a three-timescale hybrid architecture that confines the LLM to its irreplaceable role: natural language intent understanding. Our work provides an honest, empirically-grounded assessment of where LLMs add genuine value versus where simpler methods suffice.

Future work includes: (1) resolving the switching-stability dilemma through per-regime value function isolation or meta-learning initializations; (2) performance-grounded RAG anchoring~\cite{lewis2020rag, grounding2025rl} to reduce LLM weight oscillation by retrieving verified historical (KPI, weights, performance) tuples as numerical anchors; (3) multi-satellite coordination where LLMs manage inter-satellite intent negotiation; and (4) deployment on real satellite testbeds with hardware-in-the-loop validation.

\bibliographystyle{IEEEtran}
\bibliography{refs}

\begin{thebibliography}{10}
\providecommand{\url}[1]{#1}
\csname url@samestyle\endcsname
\providecommand{\newblock}{\relax}
\providecommand{\bibinfo}[2]{#2}
\providecommand{\BIBentrySTDinterwordspacing}{\spaceskip=0pt\relax}
\providecommand{\BIBentryALTinterwordstretchfactor}{4}
\providecommand{\BIBentryALTinterwordspacing}{\spaceskip=\fontdimen2\font plus
\BIBentryALTinterwordstretchfactor\fontdimen3\font minus
  \fontdimen4\font\relax}
\providecommand{\BIBforeignlanguage}[2]{{%
\expandafter\ifx\csname l@#1\endcsname\relax
\typeout{** WARNING: IEEEtran.bst: No hyphenation pattern has been}%
\typeout{** loaded for the language `#1'. Using the pattern for}%
\typeout{** the default language instead.}%
\else
\language=\csname l@#1\endcsname
\fi
#2}}
\providecommand{\BIBdecl}{\relax}
\BIBdecl

\bibitem{starlink}
{SpaceX}, ``{SpaceX Starlink},'' \url{https://www.starlink.com}, 2024,
  accessed: 2024-12-01.

\bibitem{beam_hopping}
R.~Alegre-Godoy, N.~Alagha, and M.~{\'A}. V{\'a}zquez-Castro, ``Beam hopping
  for multi-beam {GEO} satellite communication systems,'' \emph{IEEE
  Transactions on Wireless Communications}, vol.~14, no.~4, pp. 1832--1842,
  2015.

\bibitem{drl_satcom1}
X.~Hu, S.~Liu, R.~Chen, W.~Wang, and C.~Wang, ``Deep reinforcement learning for
  dynamic spectrum access in satellite communications,'' \emph{IEEE
  Transactions on Vehicular Technology}, vol.~69, no.~12, pp. 16\,135--16\,147,
  2020.

\bibitem{ma2023eureka}
Y.~J. Ma, W.~Liang, G.~Wang, D.-A. Huang, O.~Bastani, D.~Jayaraman, Y.~Zhu,
  L.~Fan, and A.~Anandkumar, ``Eureka: Human-level reward design via coding
  large language models,'' in \emph{Proc. International Conference on Learning
  Representations (ICLR)}, 2024.

\bibitem{drl_resource}
R.~Li, Z.~Zhao, Q.~Sun, I.~Chih-Lin, C.~Yang, X.~Chen, M.~Zhao, and H.~Zhang,
  ``Deep reinforcement learning for resource management in network slicing,''
  \emph{IEEE Access}, vol.~6, pp. 74\,429--74\,441, 2018.

\bibitem{luis2019drl_beam}
J.~J. Garau-Luis, M.~Guerster, I.~del Portillo, E.~Crawley, and B.~Cameron,
  ``Deep reinforcement learning architecture for continuous power allocation in
  high throughput satellites,'' in \emph{Proc. AIAA International
  Communications Satellite Systems Conference (ICSSC)}, 2019, arXiv:1906.00571.

\bibitem{ferreira2023multibeam}
P.~V.~R. Ferreira, R.~Paffenroth, A.~M. Wyglinski, T.~M. Hackett, S.~G. Bilen,
  and R.~C. Reinhart, ``Multi-objective optimization for cognitive satellite
  communications using deep reinforcement learning,'' \emph{IEEE Transactions
  on Cognitive Communications and Networking}, vol.~9, no.~4, pp. 882--897,
  2023.

\bibitem{wang2023drl_satcom_survey}
Z.~Wang, D.~Wang, Y.~Ren, and J.~Wang, ``Deep reinforcement learning for
  satellite communication: A survey,'' \emph{Space: Science \& Technology},
  vol.~3, p. 0087, 2023.

\bibitem{ng1999reward_shaping}
A.~Y. Ng, D.~Harada, and S.~Russell, ``Policy invariance under reward
  transformations: Theory and application to reward shaping,'' in \emph{Proc.
  International Conference on Machine Learning (ICML)}, 1999, pp. 278--287.

\bibitem{hayes2022morl_survey}
C.~F. Hayes, R.~R\u{a}dulescu, E.~Dechering, P.~Mannion, D.~M. Roijers,
  A.~Nowé, and P.~Vamplew, ``A practical guide to multi-objective
  reinforcement learning and planning,'' \emph{Autonomous Agents and
  Multi-Agent Systems}, vol.~36, no.~1, p.~26, 2022.

\bibitem{zheng2020reward_design}
Z.~Zheng, J.~Oh, N.~Heess, and S.~Singh, ``What can learned intrinsic rewards
  capture?'' \emph{arXiv preprint arXiv:1912.05500}, 2020.

\bibitem{llm_telecom}
L.~Bariah, N.~Naderializadeh, S.~Hashemi, and H.~Nikopour, ``Large language
  models for telecom: Opportunities and challenges,'' \emph{IEEE Communications
  Magazine}, 2024.

\bibitem{llm_network}
S.~Wang, S.~Bi, and Y.-J.~A. Zhang, ``Networking with large language models,''
  \emph{IEEE Wireless Communications}, 2024.

\bibitem{maatouk2023telecomgpt}
A.~Maatouk, N.~Piovesan, F.~Ayed, A.~De~Domenico, and M.~Debbah, ``Large
  language models for telecom: Forthcoming impact on the industry,''
  \emph{arXiv preprint arXiv:2308.06013}, 2023.

\bibitem{shao2024wirelessllm}
J.~Shao, J.~Tong, Q.~Wu, W.~Guo, Z.~Li, Z.~Lin, and J.~Zhang, ``{WirelessLLM}:
  Empowering large language models towards wireless intelligence,'' \emph{arXiv
  preprint arXiv:2405.17053}, 2024.

\bibitem{cao2024llmrl_survey}
S.~Schoepp, M.~Jafaripour, Y.~Cao, T.~Yang, F.~Abdollahi, S.~Golestan,
  Z.~Sufiyan, O.~R. Zaiane, and M.~E. Taylor, ``The evolving landscape of
  {LLM}- and {VLM}-integrated reinforcement learning,'' \emph{arXiv preprint
  arXiv:2502.15214}, 2025.

\bibitem{cusum}
E.~S. Page, ``Continuous inspection schemes,'' \emph{Biometrika}, vol.~41, no.
  1/2, pp. 100--115, 1954.

\bibitem{ppo}
J.~Schulman, F.~Wolski, P.~Dhariwal, A.~Radford, and O.~Klimov, ``Proximal
  policy optimization algorithms,'' \emph{arXiv preprint arXiv:1707.06347},
  2017.

\bibitem{abel2023continual}
D.~Abel, A.~Barreto, B.~Van~Roy, D.~Precup, H.~van Hasselt, and S.~Singh, ``A
  definition of continual reinforcement learning,'' in \emph{Advances in Neural
  Information Processing Systems}, vol.~36, 2023.

\bibitem{yu2020gradient_surgery}
T.~Yu, S.~Kumar, A.~Gupta, S.~Levine, K.~Hausman, and C.~Finn, ``Gradient
  surgery for multi-task learning,'' in \emph{Advances in Neural Information
  Processing Systems}, vol.~33, 2020.

\bibitem{lewis2020rag}
P.~Lewis, E.~Perez, A.~Piktus, F.~Petroni, V.~Karpukhin, N.~Goyal,
  H.~K\"uttler, M.~Lewis, W.-t. Yih, T.~Rockt\"aschel \emph{et~al.},
  ``Retrieval-augmented generation for knowledge-intensive {NLP} tasks,'' in
  \emph{Advances in Neural Information Processing Systems}, vol.~33, 2020.

\bibitem{grounding2025rl}
Anonymous, ``Grounding by trying: {LLMs} with {RL}-enhanced retrieval,''
  \emph{arXiv preprint arXiv:2410.23214}, 2025, iCLR 2025.

\end{thebibliography}

\end{document}